\documentclass[letterpaper]{article}
\usepackage{flairs}
\usepackage{times}
\usepackage{helvet}
\usepackage{courier}
\usepackage{graphicx}
\usepackage{amssymb}
\usepackage{latexsym}
\usepackage{url}
\usepackage{xspace}
\usepackage{ifthen}
\newcommand{\FINAL}  

\newcommand{\curricugen}{\textsc{CurricuGen}\xspace}
\newcommand{\curriculama}{\textsc{CurricuLAMA}\xspace}
\newcommand{\makeannotatedtask}{\textsc{MakeAnnotatedTask}\xspace}
\newcommand{\sfunction}[1]{\textsc{#1}}
\newcommand{\domain}{\ensuremath{\Sigma}\xspace}
\newcommand{\problem}{\ensuremath{P}\xspace}
\newcommand{\classicalproblem}{\ensuremath{P}\xspace}
\newcommand{\hierarchicalproblem}{\ensuremath{P_h}\xspace}

\newcommand{\trace}{\ensuremath{\pi}\xspace}

\newcommand{\methods}{\ensuremath{M}\xspace}
\newcommand{\curriculum}{\ensuremath{C}\xspace}

\newcommand{\learnmethod}{\sfunction{Learn-Method}\xspace}
\newcommand{\curriculearn}{\textsc{CurricuLearn}\xspace}
\newcommand{\htnmaker}{\textsc{HTN-Maker}\xspace}

\newcommand{\sv}[1]{{\normalfont\textsf{\footnotesize #1}}} 
\usepackage{algorithm}
\usepackage[noend]{algorithmic}
\usepackage{multirow}
\usepackage{newfloat}
\usepackage{listings}
\floatstyle{ruled}
\newfloat{listing}{tb}{lst}{}
\floatname{listing}{Listing}
\newcommand{\code}[1]{{\tt\footnotesize #1}}
\newcommand{\tightparagraph}[1]{\vspace{2mm}\noindent\textbf{#1}}

\usepackage[dvipsnames]{xcolor}
\newcommand{\fromruoxi}[2][]{#1\frombody{red}{Ruoxi}{#2}}
\newcommand{\frommak}[2][]{#1\frombody{blue}{mak}{#2}}

\newcommand{\frombody}[3]{
	\noindent
	\textcolor{#1}{
		{$\bf [\!\![\!\![$}\underline{\scshape{#2}}
		{\scshape says:} 
		\textsl{#3}{$\bf ]\!\!]\!\!]$}}
	{}}
\ifdefined \DRAFT
\else
  \renewcommand{\frombody}[3]{}
\fi

\usepackage[switch, modulo]{lineno}
\begin{document}
\newcommand{\includeappendix}[0]{}

\title{Automatically Learning HTN Methods from Landmarks}

\author {
  Ruoxi Li\textsuperscript{\rm 1},
  Dana Nau\textsuperscript{\rm 1},
  Mark Roberts\textsuperscript{\rm 2},
  Morgan Fine-Morris\textsuperscript{\rm 2}\\
  \textsuperscript{\rm 1}Dept. of Computer Science and Institute for Systems Research, Univ.\ of Maryland, College Park, MD, USA\\
  \textsuperscript{\rm 2}Navy Center for Applied Research in AI, Naval Research Laboratory, Washington, DC, USA\\
  rli12314@cs.umd.edu, nau@umd.edu, mark.roberts@nrl.navy.mil, morgan.fine-morris.ctr@nrl.navy.mil
}

\maketitle

\begin{abstract}
Hierarchical Task Network (HTN) planning usually requires a domain engineer to provide manual input about how to decompose a planning problem. Even \htnmaker, a well-known method-learning algorithm, requires a domain engineer to annotate the tasks with information about what to learn. We introduce \curriculama, an HTN method learning algorithm that completely automates the learning process. It uses landmark analysis to compose annotated tasks and leverages curriculum learning to order the learning of methods from simpler to more complex. This eliminates the need for manual input, resolving a core issue with \htnmaker. We prove {\curriculama}’s soundness, and show experimentally that it has a substantially similar convergence rate in learning a complete set of methods to \htnmaker.
\end{abstract}

\section{Introduction}
Automated planning systems require a domain expert to provide knowledge about the dynamics of the planning domain. In Hierarchical Task Networks (HTNs), expert-provided knowledge includes structural properties and potential hierarchical problem-solving strategies in the form of HTN decomposition methods. 
Writing these methods is a significant knowledge engineering burden.
Some techniques \cite{hogg2016learning,hogg2008htnmaker} partially overcome this burden by learning HTN methods after analyzing the semantics of a solution plan for planning problems. 
But these techniques still require input from the human designer.
We overcome the need for human input by combining two insights to produce an algorithm, \curriculama, that leverages planning landmarks and curriculum learning.
Unlike other HTN learning algorithms, \curriculama doesn't require a human to construct a curriculum because it constructs its own curriculum by analyzing the landmarks in planning problems.

Curriculum learning \cite{bengio2009curriculum} is a training strategy that improves learning performance by presenting training examples in increasing order of difficulty. 
We apply curriculum learning to the problem of learning HTN methods by emphasizing learning simpler methods before learning gradually more complex methods that incorporate previously learned methods. 

Landmarks \cite{hoffmann2004ordered} are facts that must appear in every solution to a planning problem.
In the context of learning hierarchical knowledge, methods that achieve landmarks provide a backbone for solving a planning problem.
More critically, landmarks also provide a natural way to structure methods automatically.

We develop an approach that builds curricula to learn methods that achieve landmarks.  
This paper makes the following contributions:
\begin{itemize}

\item We introduce \curriculama, which uses landmarks to generate curricula for constructing HTN methods.
This approach obviates \htnmaker's need for manual annotation of tasks. \fromruoxi{refined this}

\item We prove that the methods learned by \curriculama can be used by a hierarchical planner to solve an HTN planning problem that is equivalent to the classical planning problem from which the methods were learned.

\item
Our experimental results show that \curriculama has a similar convergence rate to \htnmaker in learning a complete set of methods to solve all the test problems.

\end{itemize}

\section{Background}

\tightparagraph{\htnmaker.}
Our work builds on the HTN method learning mechanism of the \htnmaker algorithm, which learns hierarchical planning knowledge in the form of decomposition methods for HTNs. \htnmaker takes as input the initial states from a set of classical planning problems in a planning domain and solutions to those problems, as well as a set of semantically-annotated tasks to be accomplished. The algorithm analyzes this semantic information in order to determine which portions of the input plans accomplish a particular annotated task and constructs HTN methods based on those analyses.
Formally, each annotated task has a head, preconditions, and goals.  For example, here is the Make-Clear annotated task for the Blocks World domain:\smallskip
    \begin{lstlisting}[mathescape=true]
    (:task
     :head Make-Clear(?a - block)
     :preconditions ()
     :goals ((clear ?a)))
    \end{lstlisting}
What we call goals \htnmaker calls effects. It does not cause the goals to be true; instead, the goals specify what is needed to be true after performing the annotated task.

\newcommand{\firstadded}[1]{\ensuremath{#1'}}
\newcommand{\leftlm}{\ensuremath{l_i}\xspace}
\newcommand{\rightlm}{\ensuremath{l_j}\xspace}
\newcommand{\firstleftlm}{\firstadded{l_i}\xspace}
\newcommand{\firstrigthlm}{\firstadded{l_j}\xspace}
\tightparagraph{Landmarks.}
\emph{Landmarks} are a natural way to subdivide a planning solution.
A landmark \cite{hoffmann2004ordered} for a planning problem is a fact that is true at some point in every plan that solves the problem. A landmark graph is a directed graph where the nodes are landmarks and the edges are orderings. There are four types of orderings among landmarks: 
natural, necessary, greedy-necessary, and reasonable
\ifthenelse
{\isundefined{\includeappendix}}
{(see Appendix for formal definitions\footnote{A version of this paper with an appendix is available on ArXiv. \fromruoxi{TO-DO: insert link}}).}
{(see Appendix \ref{sec:landmarks_appendix} for formal definitions.).}

\section{Learning HTN methods with Curricula Generated from Landmarks}

Suppose we want to teach an HTN learner to learn methods for solving some task $\tau$. An ideal curriculum would focus the learner on the simplest subtasks of $\tau$ first, then build more and more complex subtasks until all of $\tau$ is learned. 
Learning HTN methods from plan traces is a common approach, and it follows that traces for subtasks of $\tau$ will also be subtraces of a plan trace for $\tau$. Specifically,
if $\trace$ is a plan trace for solving $\tau$, then the plan trace for solving each subtask $\tau$ is a subtrace $\trace[b,e]$ of $\trace$, where $[b, e]$ indicates the beginning and ending indices of the subtrace.
Thus we can represent a $k$-step curriculum that learns subtraces as a sequence of triples of the form $(b_i, e_i,\tau_i)$ where $i \in \mathbb{N}_k$. \fromruoxi{modified here}

\tightparagraph{Definition 1.} 
A \emph{classical planning problem} 
\problem is a triple 
$(\Sigma, s_0, g)$, 
where $\Sigma$ is the classical planning domain description, $s_0$ is the initial state and $g$ is the goal \cite{ghallab2016automated}.

\tightparagraph{Definition 2.} Given a plan trace \trace, a \emph{curriculum} $C$ is a sequence of \textit{k} curriculum steps of the form $(b_i, e_i, \tau_i)$, where $b_i$ and $e_i$ specify the starting and ending indices of the subtrace to analyze, and $\tau_i$ specifies the annotated task to learn from the subtrace for step $i \in \mathbb{N}_k$. \fromruoxi{modified here}

\vspace{2mm}

\begin{figure}[t]
\begin{minipage}{.32\columnwidth}
\centering
    \includegraphics[width=.6\linewidth]{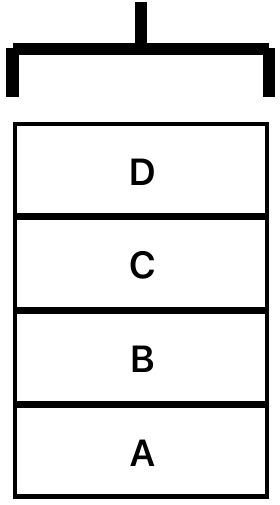}
\end{minipage}
\hspace{\fill}
\begin{minipage}{.68\columnwidth}
    \begin{center}\small
        \begin{tabular}{l}
        Action 1: \code{(!Unstack D C)} \\
        Action 2: \code{(!Putdown D)} \\
        Action 3: \code{(!Unstack C B)} \\
        Action 4: \code{(!Putdown C)} \\
        Action 5: \code{(!Unstack B A)} 
        \end{tabular}
    \end{center}
\end{minipage}

\caption{A Blocks World problem in which the initial state is a stack of 4 blocks. The goal is to make the bottom block A clear. The plan to achieve the goal is shown on the right.}
\label{fig:blocks_demo}

\end{figure}

For example, consider a Blocks World classical planning problem with 4 blocks A, B C and D stacked on each other (see Figure \ref{fig:blocks_demo}) and the goal is to have block A's top clear. Formally: 
initial state $s_0$=\{(on-table A), (on B A), (on C B), (on D C), (clear D), (hand-empty)\}; and goal $g$=\{(clear A)\}.
A possible solution $\trace_{clearA}$ is to remove the blocks above A one by one through 5 actions, which are prefixed with the character `!' to distinguish them from predicates.
Here is a
3-step curriculum
for learning methods of $\trace_{clearA}$:
\begin{center}\small
\begin{tabular}{c|ccc}
Step & Begin & End & Annotated Task\\
\hline
$a$ & 1 & 1 & \code{Make-Clear}\\
$b$ & 1 & 3 & \code{Make-Clear}\\
$c$ & 1 & 5 & \code{Make-Clear}\\
\end{tabular}
\end{center}

We want step $a$ to learn a method \textit{m1} for annotated tasks \code{Make-Clear} from the first action in plan $P$, this method tries to clear a block under one block. Then we want step $b$ to learn a method \textit{m2} for annotated tasks \code{Make-Clear} from the first to the third action in plan $P$, this method tries to clear a block under two blocks, and would presumably contain a subtask \code{Make-Clear} that is related to \textit{m1} previously learned from the first action. Then step $c$ learns a new method \textit{m3} for \code{Make-Clear} that subsumes \textit{m2}.

\curriculama, has two subroutines: \curricugen generates curricula from landmarks while \curriculearn  learns HTN methods from the curricula. 

\subsection{\curricugen}

Since a landmark must be true at some point in every solution to a planning problem, we hypothesized that it would be useful to learn methods that reach landmarks. Our algorithm, \curricugen
(Algorithm \ref{alg:c-lama}) extracts landmarks from a planning problem, then generates a curriculum from those landmarks.

\begin{algorithm}[t]
\caption{\curricugen: Curriculum Generation from Landmarks}
\label{alg:c-lama}
\textbf{Input}: a classical planning problem \classicalproblem,  a possibly empty set of HTN
methods $M$ \\
\textbf{Output}:  an updated set of HTN methods $M$

\begin{algorithmic}[1]
\STATE $(\Sigma, s_0, g) \leftarrow P$
\STATE $(V, E_V)$ $\leftarrow$ extract landmark graph from $(\Sigma, s_0, g)$ \label{line:hm} 
\STATE add reasonable orders to $(V, E_V)$ \label{line:hps} 
\STATE $C \leftarrow \langle \rangle$ \COMMENT{initialize the curriculum steps}
\STATE $\pi \leftarrow \langle \rangle$ \COMMENT{initialize the plan trace}
\STATE $s \leftarrow s_0$ \COMMENT{initialize the current state}
\STATE $i \leftarrow 0$ \COMMENT{initialize the plan length}
\WHILE{$V \neq \emptyset$}
    \STATE select and remove a vertex $v$ in $(V, E_V)$ that has no predecessors
    \STATE $\pi' \leftarrow \sfunction{ClassicalPlanner}(\Sigma, s, v)$ \label{line:plan}
    \STATE $s \leftarrow \gamma(s, \pi)$ \label{line:update}
    \STATE $i \leftarrow i + length(\pi')$
    \STATE concatenate $\pi'$ to $\pi$
    \STATE $t \leftarrow \makeannotatedtask(v)$ \label{line:maketask}
    \FOR{$k$ from $i$ to 1} \label{line:addcurriculumstepI}
        \STATE append $(k, i, t)$ to $C$ \label{line:addcurriculumstepII}
    \ENDFOR
\ENDWHILE
\STATE $M$ = $\curriculearn(\Sigma, s_0, \trace, \curriculum, M)$ \label{line:htnlearn}
\STATE \textbf{return} $M$
\end{algorithmic}
\end{algorithm}

\curricugen takes as input a classical planning problem. It first generates a landmark graph for $P$ using $h^m$ Landmarks (Line \ref{line:hm}) from the landmark generation method introduced by \citeauthor{hmlandmarks} (\citeyear{hmlandmarks}); 
then it adds reasonable orders to the landmark graph (Line \ref{line:hps}) from the method described by \citeauthor{hoffmann2004ordered} (\citeyear{hoffmann2004ordered});\footnote{We use the implementation of $h^m$ landmark generation and reasonable order extraction in the Fast Downward planning system (https://www.fast-downward.org/), configured to only allow for single-atom (conjunctive) landmarks. }
and then it iterates through the landmarks by their orderings. 
For each landmark, \curricugen iteratively obtains a solution trace from the current state using a classical planner and updates the current state by applying the solution plan (Lines \ref{line:plan} and \ref{line:update}). \makeannotatedtask (Line \ref{line:maketask}) takes as input the current landmark and produces an annotated task that has a task name, empty preconditions, and the landmark as its goals. 
Given the annotated task produced from the landmark, \curricugen generates curriculum steps (Lines \ref{line:addcurriculumstepI} and \ref{line:addcurriculumstepII})
that progressively trace backward to the beginning of the plan to learn methods.

\tightparagraph{Example.}
\begin{figure}[t]
\centering
\includegraphics[width=\columnwidth]{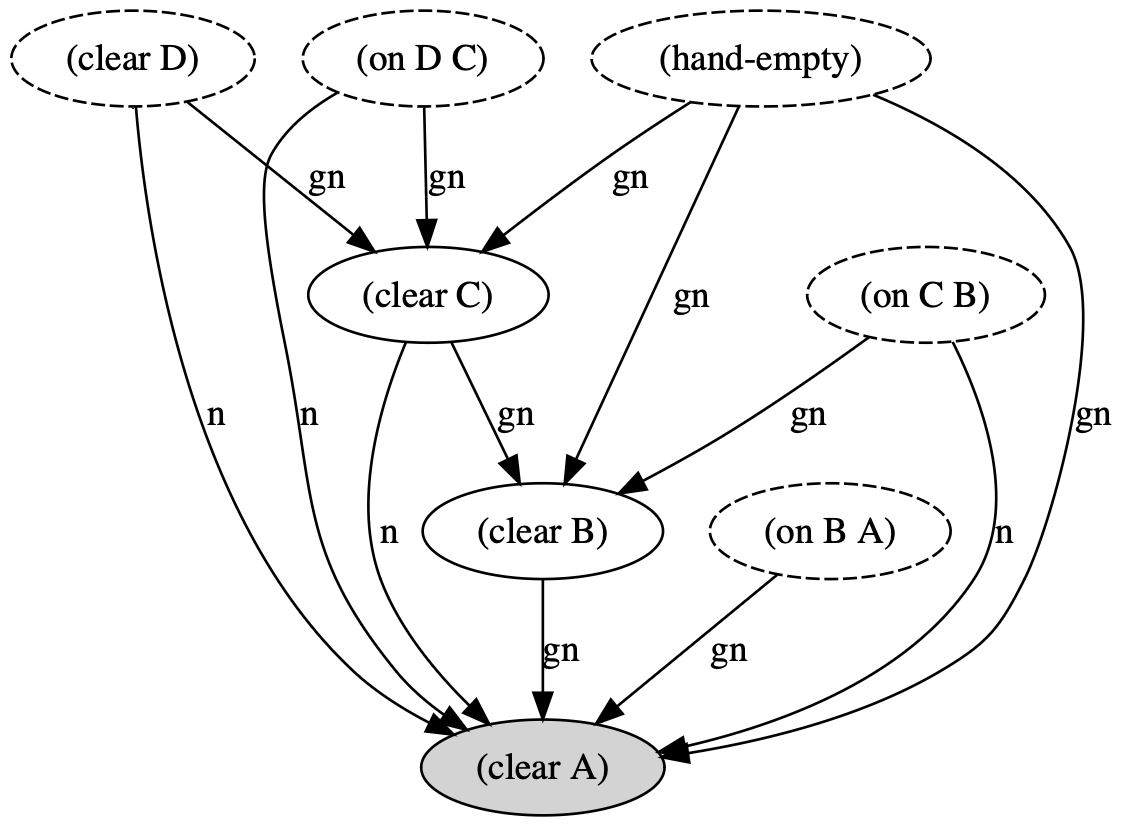}
\caption{A landmark graph for clearing block A from blocks B, C and D above in the Blocks World domain. The circled nodes are landmarks, where the dashed nodes are the landmarks that are satisfied in the initial state, and the filled node is the goal. The edges are orderings among the landmarks, where `gn' stands for greedy necessary ordering, and `n' stands for natural ordering.}
\label{fig:lm_graph_blocks}
\end{figure}
In the Blocks World problem in Figure \ref{fig:blocks_demo},
the initial state has 4 stacked blocks \textsf{A}, \textsf{B}, \textsf{C}, and \textsf{D}, and the goal is to clear block \textsf{A}.
Excluding the initial state, Figure \ref{fig:lm_graph_blocks} shows the landmark graph from \curricugen for $(s_0,g)$, which consists of 3 landmarks: $\sv{(clear C)} \prec \sv{(clear B)} \prec \sv{(clear A)}$.

\begin{figure}[t]
\includegraphics[width=1.01\columnwidth]{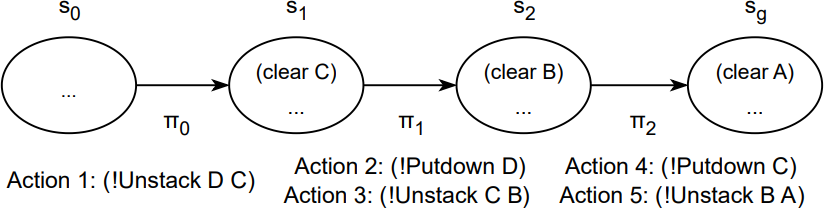}
\caption{The subplans generated from the landmarks. 
}
\label{fig:curriculumdemo}
\end{figure}

\curricugen generates subplans to achieve the first, second and the third landmarks (Figure \ref{fig:curriculumdemo}). For each landmark, it creates an annotated task (in this case, \code{Make-Clear}) and curriculum steps in which the final indices correspond to the action that achieves the landmark, and the beginning indices go back to the plan trace's start:
\begin{center}\small
\begin{tabular}{c|ccc}
Step & Begin & End & Annotated Task\\
\hline

a & 1 & 1 & \code{Make-Clear}\\
b & 3 & 3 & \code{Make-Clear}\\
c & 2 & 3 & \code{Make-Clear}\\
d & 1 & 3 & \code{Make-Clear}\\
e & 5 & 5 & \code{Make-Clear}\\
f & 4 & 5 & \code{Make-Clear}\\
g & 3 & 5 & \code{Make-Clear}\\
h & 2 & 5 & \code{Make-Clear}\\
i & 1 & 5 & \code{Make-Clear}\\

\end{tabular}
\end{center}

The curriculum comprises nine steps, labeled from \textit{a} to \textit{i}. Each step is defined by a specific segment of \trace, delineated by its beginning and ending indices, along with the name of an annotated task. The curriculum is structured to initiate simpler tasks, gradually progressing to more complex ones. The regressive sequencing of the beginning indices aims at learning methods with varying preconditions for the same annotated tasks.

\subsection{\curriculearn} \label{sec:teachableHTNMaker}

\curriculama's 
\curriculearn subroutine learns HTN methods from curricula, shown in Algorithm~\ref{alg:c-learn}.
The input to \curriculearn includes a domain description \domain, an initial states $s_0$, an execution trace \trace (which can be a plan produced by a planner given a goal), a curriculum \curriculum, and a possibly empty set of HTN methods $M$.
For each curriculum step in \curriculum, it uses the algorithm \learnmethod 
\footnote{\learnmethod performs hierarchical goal regression over a plan trace. It is the same procedure that \htnmaker uses as a subroutine \cite[Algorithm 3]{hogg2016learning} to learn preconditions and subtasks of HTN methods.}
to perform the analysis on the subtrace $\trace[b ,e]$ and learns some new methods for $\tau$. It also keeps a set of indexed method instances $X$ to identify and reuse previously learned methods as subroutines in a new method that is being synthesized. Specifically, each method is indexed by the beginning index of the corresponding subtrace $b$, and the ending index of the corresponding subtrace $e$.

\begin{algorithm}[ht]
\caption{A high-level description of \curriculearn.}
\textbf{Input}: classical domain description $\Sigma$, initial state $s_0$, solution trace $\pi$, curriculum $C$, and a possibly empty set of HTN methods $M$\\
\textbf{Output}: an updated set of HTN methods

\label{alg:c-learn}
\begin{algorithmic}[1] 

\STATE initialize $X \leftarrow \emptyset$
\STATE let $\vec{S}$ be the state trajectory generated from $\gamma(s_{0}, \pi)$
\FOR{$(b,e,\tau) \in C$} \label{line:curriculum}
    \STATE $M' \leftarrow \learnmethod (\pi, \vec{S}, \tau, X, b, e)$ \label{line:learnmethod}
    \STATE $M \leftarrow M \cup M'$
    \FOR{$m \in M'$}
        \STATE $X \leftarrow X$ $\cup$ \{($m$, $b$, $e$)\} \label{line:method}
    \ENDFOR
\ENDFOR
\STATE \textbf{return} $M$
\end{algorithmic}
\end{algorithm}


%

\curriculama takes as input a planning problem and outputs a set of learned HTN methods. It does this by using \curricugen to generate curricula from landmarks, and \curriculearn to acquire HTN methods from these curricula. 
This obviates \htnmaker's need for manual annotation of tasks and corresponding plan subtraces.

\section{Theoretical Analysis}

We prove that the methods learned by \curriculama from a classical planning problem can be applied to solve the equivalent hierarchical problem.
First, we need to go through the process of \curricugen given a classical planning problem and define the hierarchical planning problem.

Given a classical planning problem $\problem$ = $(\Sigma, s_{0}, g)$ as a training example, \curricugen produces a solution trace \trace and a curriculum \curriculum. 
Given \trace, \curriculum, and a possibly empty set of HTN methods \methods, \curriculearn will add newly learned methods to \methods.
Let $\tau$ be an annotated task that has $g$ as its goals. 
Then \hierarchicalproblem = $((\Sigma, M), s_0, \langle\tau\rangle)$ is the \textit{hierarchical planning problem} (see Definition 3) equivalent to the classical planning problem \problem.

\tightparagraph{Definition 3.} A \emph{hierarchical planning problem} \hierarchicalproblem is a triple $(\Sigma_h, s_0, \langle\tau\rangle)$ where $\Sigma_h$ is the hierarchical planning domain description, $s_0$ is the initial state and $\langle\tau\rangle$ is the task list. A hierarchical planning domain description $\Sigma_h$ is a tuple $(\Sigma, M)$, where $\Sigma$ is the classical planning domain description and $M$ is the set of HTN methods. 

\vspace{2mm}
Now we can show how the methods learned by \curriculama can be used to solve the hierarchical problem, which is equivalent to the classical problem from which the methods were learned.

\tightparagraph{Proposition 1.}
\emph{
Given \problem = $(\Sigma, s_{0}, g)$, \trace, $\tau$, M, and \hierarchicalproblem = $((\Sigma, M), s_0, \langle\tau\rangle)$,
\trace is a solution to \hierarchicalproblem as a result of hierarchically decomposing $g$ using the methods in $M$.
}\vspace{1ex}

\textbf{Proof Sketch} If \trace is empty or $g$ is satisfied in $s_0$, then \curriculearn will learn a trivial method for $\tau$ that has empty subtasks, which is sufficient to solve the problem. Otherwise, since $g$ has to be the final landmark in the landmark graph of $P$, the final curriculum step in $C$ is $(1, len(\pi), \tau)$. Therefore, \curriculearn will learn at least one method from curriculum step $(1, len(\pi), \tau)$. This method must be applicable to $s_0$ in \hierarchicalproblem because its preconditions were computed by regressing g through the actions of \trace \cite[the \learnmethod precedure]{hogg2016learning}, which is applicable to $s_0$. Furthermore, the goal regression procedure guarantees that whenever the preconditions of a method are satisfied there must be some way to reduce the subtasks of that method using other methods learned from \trace,
because the subtasks of that method were chosen from the indexed instances of other methods. Eventually, $g$ in \hierarchicalproblem will be hierarchically decomposed into the action sequence \trace.
$\square$
\vspace{2mm}

Therefore, \curriculama is sound for the original problem for which it learned methods.  That is,  methods learned by \curriculama from a classical planning problem \problem will solve the equivalent hierarchical problem \hierarchicalproblem. 
However, we also want to know how rapidly it can learn a complete set of methods from the training problems.  
In the next section, we will empirically evaluate \curriculama to show that it has a comparable convergence rate in learning a complete set of methods to \htnmaker.

\begin{figure*}[ht]
    \centering
    \includegraphics[width=\textwidth]{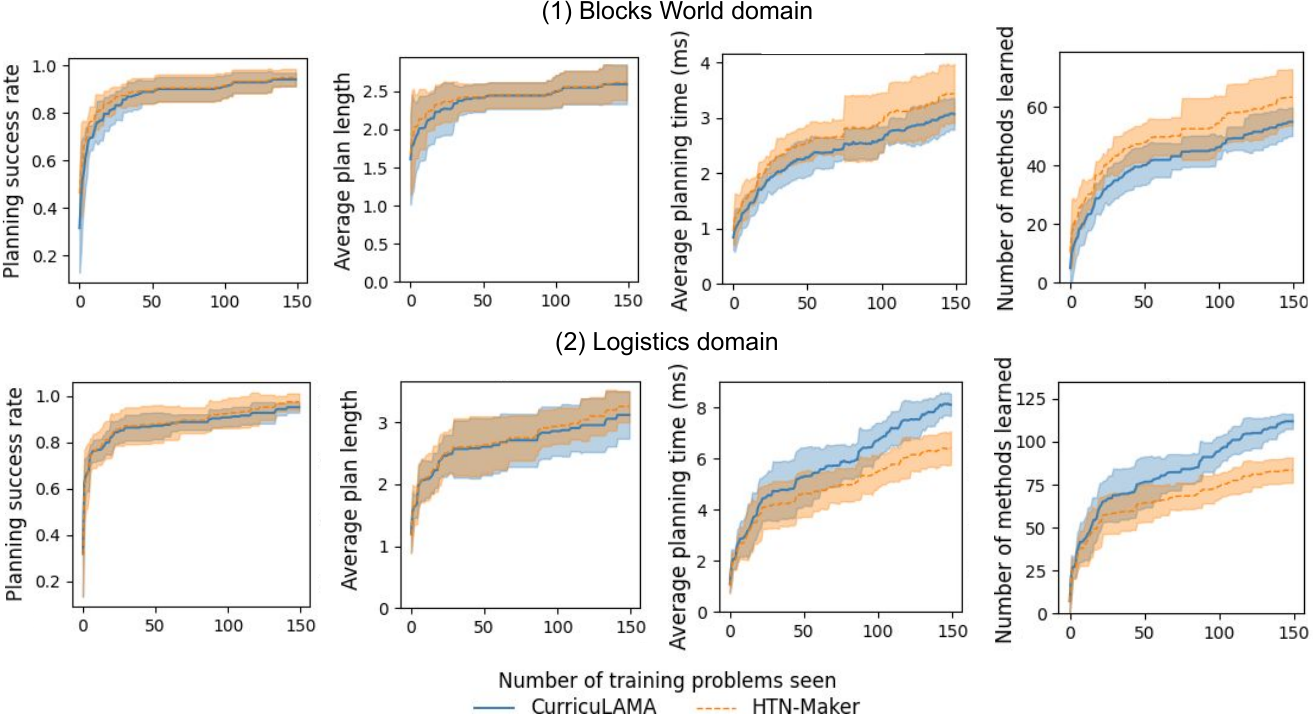}
    \caption{
    \emph{Experimental results} in (1) the Blocks World domain and (2) the Logistics domain.
        From left to right, the subfigure's y-axis shows (a) the fraction of problems that the planner could successfully solve using the methods that each learning algorithm learned; (b)  the average length of the plans that the planner produced using the learned methods;
        (c) the average planning time over the 50 test problems;
        and (d) the total number of methods learned.
        In each of the subfigures, the x-axis shows the number of training problems (0-150) from which the methods were learned.  The blue line displays the results for \curriculama and the orange dashed line displays the results for \htnmaker. The shaded areas indicate the variance in the number of methods learned across five trials. 
    }
    \label{fig:blocks_results}
\end{figure*}


\section{Empirical Study}

We have evaluated \curriculama (and compared it to \htnmaker) experimentally in five IPC (International Planning Competition) domains: Logistics, Blocks World, Rover, Satellite, and Zeno Travel.
These domains are used for evaluation in the original papers on HTN-Maker.
Due to space limitations, we present results for the Blocks World and Logistics; results are similar across domains and can be seen in 
\ifthenelse{\isundefined{\includeappendix}}
{the Appendix.}
{Appendix \ref{sec:domain_description_appendix} and \ref{sec:appendix:more_exp}.}
We assess the efficiency of \curriculama in learning Hierarchical Task Network methods and the effectiveness in solving hierarchical planning problems using the learned methods. 

Our evaluation considered how well the methods learned from an incremental set of training problems can solve a set of static testing problems. A single trial for a domain used PDDL-Generators \cite{seipp-et-al-zenodo2022} to generate 150 random training problems and 50 testing problems from the same distribution of parameters.  Starting with an empty set of methods $M$, the procedure iterated through the training problems $(1, 2, .., 150)$, augmented $M$ using either \curriculama or \htnmaker, and used \htnmaker's version of the SHOP planner \cite{SHOP} to solve the 50 test problems with the current set $M$. 
We repeated five trials in each of the five domains and reported on the following metrics: (1) convergence, (2) average plan length, (3) average planning time, (4) method generation, (5) running time in learning. All experiments are run on AMD EPYC 7763 (2.45 GHz)\fromruoxi{added this info}.

Figure \ref{fig:blocks_results} shows that \curriculama's method learning exhibits a similar convergence rate and results in plan lengths and planning time comparable to \htnmaker, all while achieving a significant advantage: it \emph{completely eliminates} the need for expert-provided annotated tasks. 

\curriculama's planning mechanism may cause it to learn extraneous methods in some domains (e.g., the Logistics, Satellite and Rover domain\fromruoxi{added Satellite, just to point out methods are extraneous in those domains}). 
While it's possible that this may be an indication of overfitting, we believe this is more likely a result of partial orders in the landmark graph.
The landmark generation algorithms used by \curriculama (Algorithm \ref{alg:c-lama}, lines \ref{line:hm} and \ref{line:hps}) return only a partial ordering among landmarks given the additional reasonable orders. All reasonable orderings are not determined because determining whether a reasonable order exists between two given landmarks is a PSPACE-complete problem \cite{hoffmann2004ordered}. Then, \curriculama enforces a total ordering to formulate a sequence of subgoals, which is not necessarily the optimal strategy
\ifthenelse
{\isundefined{\includeappendix}}
{(see Appendix for an example).}
{(see Appendix \ref{sec:illustration_appendix} for an example).}
This often results in \curriculama's derivation of additional methods from extended plan traces, as those methods cover more potential (and sub-optimal) paths to the goal.
Improving \curriculama's strategy for ordering landmarks by incorporating more sophisticated heuristics could help reduce the creation of these redundant methods and is a topic for future work.

While \curriculama may learn slightly more methods in some domains due to suboptimal landmark orderings, this does not appear to have a detrimental impact on planning success rates or plan lengths. 
Notably, in domains where \curriculama successfully captures all required landmark orderings, it learns fewer methods than \htnmaker, which results in relatively shorter planning time.

\begin{figure}[ht]
\centering
\includegraphics[width=\columnwidth]{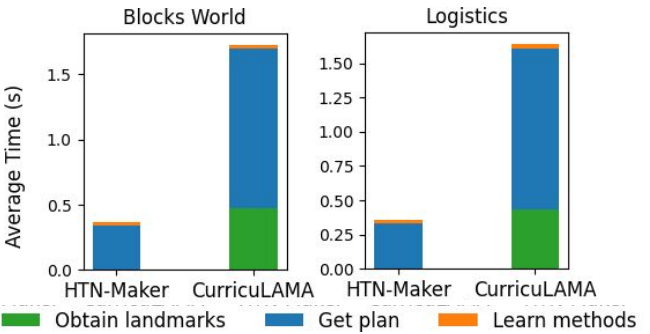}
\caption{\emph{Running time needed to learn methods.}
 The bars represent the average time that each learning algorithm spent on different parts of the learning process. Green represents the time to obtain landmarks (Alg \ref{alg:c-lama}, Line 2 and 3), blue indicates the time to obtain the plan (Alg \ref{alg:c-lama}, Line 8 to 16)
 \fromruoxi{added the reference to \curriculama}
 , and orange shows the time to learn methods.
 }
\label{fig:block_algorithm_times}
\end{figure}

It's also worth highlighting that \curriculama's additional computational time (Figure \ref{fig:block_algorithm_times}), averaging between 0.2 to 0.8 seconds per problem, is negligible when compared to the overall planning time or the time required from a domain expert. Thus, \curriculama is an efficient and promising alternative to acquiring planning methods.

\subsection{Future Work}
\fromruoxi{added this section}
The number of methods and planning time keep increasing without convergence for both algorithms in the Blocks World, Logistics and Rover domain. We believe that given enough training problems, they will eventually converge. To verify that, we will expand the training set in our future experiments.

We are also interested in an empirical study that compares manually annotated task and automatically annotated tasks when directly applied to \htnmaker without any curricula. This will give an idea of the quality of the tasks generated by \curricugen.

Last but not least, we will theoretically and empirically analyse \curriculama's time complexity as some measure of task domain problem or solution complexity increases.


\section{Related work}

Several researchers have investigated ways to learn structural knowledge, including HTNs,
though they did not use curricula generated from landmarks to guide learning.

The algorithm by \citeauthor{lotinac2016constructing}~(\citeyear{lotinac2016constructing}) uses invariant analysis to construct an HTN from the PDDL description of a planning domain and a single representative instance.
Learn-HTN \cite{zhuo2009learning} learns HTN-method preconditions and action preconditions and effects under partial observability. It receives task decomposition trees whose leaves are all primitive actions. It solves constraints using MAX-SAT among subtask-pairs in a task decomposition tree to learn action models and method preconditions. Similar to Learn-HTN, HTNLearn \cite{zhuo2014learning} learns HTN methods and action models from partially observed plan traces annotated with potentially empty partial decomposition trees that capture task-subtask decompositions.

The HTN Learning System \cite{yang2007learning} uses partitioning to learn HTN methods. Other algorithms learn method preconditions using the hierarchical relationships between tasks, the action models, a complete description of the intermediate states using case-based reasoning \cite{xu2005domain} and version-space learning \cite{ilghami2005learning}.
The algorithm by \citeauthor{li2010learning}~(\citeyear{li2010learning}) uses techniques similar to probabilistic context-free grammar learning and learns probabilistic HTNs. It can learn recursive methods on repeated structures. 

Word2HTN \cite{gopalakrishnanword2htn} combines semantic text analysis techniques and subgoal learning to create HTNs. Plan traces are viewed as sentences where a plan trace consists of actions with their grounded preconditions and effects. Each word in the sentence is an atom or an action in the plan trace. This work uses Word2Vec to convert each word into a vector representation and applies agglomerative hierarchical partitioning on the learned vectors to learn methods with binary-subtask decompositions. As an extension to their approach, \cite{fine2020learning} approximates landmarks using solution traces and learns methods with symbolic and numeric preconditions that initially decompose problems using two or more landmarks, and then finish the decomposing using (primitive task, complex task) right-recursion.

CC-HTN \cite{hayes2016AutonomouslyConstructingHierarchical} and Circuit-HTN \cite{chen2021LearningHierarchicalTask} translate execution traces into multi-trace graphical representations where primitive tasks comprise vertices and edges indicate sequential tasks. They apply bottom-up consolidation techniques to group simple tasks into progressively larger ones. Circuit-HTN treats the graph as a circuit of resistors, using techniques from the reduction of parallel and in-series resistors to discover new decompositions. CC-HTN similarly uses clique- and chain-detection for structure-learning and also learns the method parameters.

\citeauthor{herail2023LeveragingDemonstrationsLearning}~(\citeyear{herail2023LeveragingDemonstrationsLearning}) iteratively learns HTNs using bottom-up pattern-recognition and compression techniques on common sequences of subtasks in traces, which 
get replaced with a task symbol.
\citeauthor{segura2017learning}~(\citeyear{segura2017learning}) learn HTNs using process and data mining by converting execution traces into event-logs and extracting the preconditions and effects for each task using a fuzzy rule-based learning algorithm.

HPNL \cite{langleylearning} is a system that learns new methods for  Hierarchical Problem Networks \cite{shrobehierarchical} by analyzing sample hierarchical plans, using violated constraints to identify state conditions, and ordering conflicts to determine goal conditions. 


The hierarchical plan recognition algorithm by
\citeauthor{geib2009delaying}~(\citeyear{geib2009delaying})
uses Combinatory Categorial Grammars (CCGs) as part of the ELEXIR framework. It requires a hand-authored CCG representing structure of plans done by agents. Lex$_{Learn}$ 
\cite{geib2018learning} learns CCGs by enumerating CCG categories for a set of plan traces from templates. Lex$_{Greedy}$ \cite{kantharaju2021learning} employs a greedy approach to improve the scalability of CCG learning. Lex$^T_{Greedy}$ \cite{kantharaju2021learning} learns CCGs in domains with type trees as an extension to Lex$_{Greedy}$.

Teleoreactive Logic Programs (TLPs) are a framework for encoding knowledge using ideas from logic programming, reactive control, and HTNs. 
This work includes ways to learn recursive TLPs from problem solution traces \cite{choi2005learning},
a learning method that acquires recursive forms of TLP structures from traces of successful problem solving \cite{langley2006learning},
and an incremental learning algorithm for TLPs based on explanation-based learning  \cite{nejati2006learning}.


\section{Conclusion}

\curriculama generates curricula from
landmarks and uses them to acquire HTN methods according to these curricula. 
We have proved that the methods that \curriculama learns from classical planning problems enable an HTN planner to solve equivalent HTN planning problems. 
In our experiments \curriculama learned comparably good
methods to those learned by \htnmaker for the same problems, with no requirement for a human to provide methods, curricula, or annotations of tasks.
The idea that landmarks are useful for structural knowledge learning, and that curricula can be constructed from those landmarks, may apply to other structural knowledge learning techniques.


\section*{Acknowledgements}
The authors thank the reviewers, whose comments helped to improve the paper.
This work has been supported for UMD
in part by ONR grant N000142012257 and AFRL contract FA8750-23-C-0515. MR and MFM thank ONR and NRL
for funding this research.
The views expressed are those of the authors and do not reflect the official policy or position of the funders.

\clearpage
\bibliography{flairs}

\begin{thebibliography}{}

\bibitem[\protect\citeauthoryear{Bengio \bgroup et al\mbox.\egroup }{2009}]{bengio2009curriculum}
Bengio, Y.; Louradour, J.; Collobert, R.; and Weston, J.
\newblock 2009.
\newblock Curriculum learning.
\newblock In {\em ICML},  41--48.

\bibitem[\protect\citeauthoryear{Chen \bgroup et al\mbox.\egroup }{2021}]{chen2021LearningHierarchicalTask}
Chen, K.; Srikanth, N.~S.; Kent, D.; Ravichandar, H.; and Chernova, S.
\newblock 2021.
\newblock Learning {{Hierarchical Task Networks}} with {{Preferences}} from {{Unannotated Demonstrations}}.
\newblock In {\em CoRL},  1572--1581.

\bibitem[\protect\citeauthoryear{Choi and Langley}{2005}]{choi2005learning}
Choi, D., and Langley, P.
\newblock 2005.
\newblock Learning teleoreactive logic programs from problem solving.
\newblock In {\em ILP},  51--68.

\bibitem[\protect\citeauthoryear{Fine-Morris \bgroup et al\mbox.\egroup }{2020}]{fine2020learning}
Fine-Morris, M.; Auslander, B.; Floyd, M.~W.; Pennisi, G.; Mu{\~n}oz-Avila, H.; and Gupta, S. K.~M.
\newblock 2020.
\newblock Learning hierarchical task networks with landmarks and numeric fluents by combining symbolic and numeric regression.
\newblock In {\em ACS}.

\bibitem[\protect\citeauthoryear{Geib and Kantharaju}{2018}]{geib2018learning}
Geib, C., and Kantharaju, P.
\newblock 2018.
\newblock Learning combinatory categorial grammars for plan recognition.
\newblock In {\em AAAI},  3007–3014.

\bibitem[\protect\citeauthoryear{Geib}{2009}]{geib2009delaying}
Geib, C.~W.
\newblock 2009.
\newblock Delaying commitment in plan recognition using combinatory categorial grammars.
\newblock In {\em IJCAI},  1702–1707.

\bibitem[\protect\citeauthoryear{Ghallab, Nau, and Traverso}{2016}]{ghallab2016automated}
Ghallab, M.; Nau, D.; and Traverso, P.
\newblock 2016.
\newblock {\em Automated planning and acting}.
\newblock Cambridge University Press.

\bibitem[\protect\citeauthoryear{Gopalakrishnan, Munoz-Avila, and Kuter}{2016}]{gopalakrishnanword2htn}
Gopalakrishnan, S.; Munoz-Avila, H.; and Kuter, U.
\newblock 2016.
\newblock Word2{HTN}: Learning task hierarchies using statistical semantics and goal reasoning.
\newblock In {\em IJCAI Workshop on Goal Reasoning}.

\bibitem[\protect\citeauthoryear{Hayes and Scassellati}{2016}]{hayes2016AutonomouslyConstructingHierarchical}
Hayes, B., and Scassellati, B.
\newblock 2016.
\newblock Autonomously constructing hierarchical task networks for planning and human-robot collaboration.
\newblock In {\em ICRA},  5469--5476.

\bibitem[\protect\citeauthoryear{Hoffmann, Porteous, and Sebastia}{2004}]{hoffmann2004ordered}
Hoffmann, J.; Porteous, J.; and Sebastia, L.
\newblock 2004.
\newblock Ordered landmarks in planning.
\newblock {\em JAIR} 22:215--278.

\bibitem[\protect\citeauthoryear{Hogg, Mu\~{n}oz Avila, and Kuter}{2008}]{hogg2008htnmaker}
Hogg, C.; Mu\~{n}oz Avila, H.; and Kuter, U.
\newblock 2008.
\newblock {HTN-MAKER}: Learning {HTNs} with minimal additional knowledge engineering required.
\newblock In {\em AAAI},  950–956.

\bibitem[\protect\citeauthoryear{Hogg, Mu{\~n}oz-Avila, and Kuter}{2016}]{hogg2016learning}
Hogg, C.; Mu{\~n}oz-Avila, H.; and Kuter, U.
\newblock 2016.
\newblock Learning hierarchical task models from input traces.
\newblock {\em Computational Intelligence} 32(1):3--48.

\bibitem[\protect\citeauthoryear{Hérail and Bit-Monnot}{2023}]{herail2023LeveragingDemonstrationsLearning}
Hérail, P., and Bit-Monnot, A.
\newblock 2023.
\newblock Leveraging {{Demonstrations}} for {{Learning}} the {{Structure}} and {{Parameters}} of {{Hierarchical Task Networks}}.
\newblock In {\em FLAIRS}, volume~36.

\bibitem[\protect\citeauthoryear{Ilghami \bgroup et al\mbox.\egroup }{2005}]{ilghami2005learning}
Ilghami, O.; Munoz-Avila, H.; Nau, D.~S.; and Aha, D.~W.
\newblock 2005.
\newblock Learning approximate preconditions for methods in hierarchical plans.
\newblock In {\em ICML},  337--344.

\bibitem[\protect\citeauthoryear{Kantharaju}{2021}]{kantharaju2021learning}
Kantharaju, P.
\newblock 2021.
\newblock {\em Learning Decomposition Models for Hierarchical Planning and Plan Recognition}.
\newblock Ph.D. Dissertation, Drexel University, Philadelphia, Pennsylvania, USA.

\bibitem[\protect\citeauthoryear{Keyder, Richter, and Helmert}{2010}]{hmlandmarks}
Keyder, E.; Richter, S.; and Helmert, M.
\newblock 2010.
\newblock Sound and complete landmarks for and/or graphs.
\newblock In {\em ECAI},  335–340.

\bibitem[\protect\citeauthoryear{Langley \bgroup et al\mbox.\egroup }{2006}]{langley2006learning}
Langley, P.; Choi, D.; Olsson, R.; and Schmid, U.
\newblock 2006.
\newblock Learning recursive control programs from problem solving.
\newblock {\em Journal of Machine Learning Research} 7(3).

\bibitem[\protect\citeauthoryear{Langley}{2022}]{langleylearning}
Langley, P.
\newblock 2022.
\newblock Learning hierarchical problem networks for knowledge-based planning.
\newblock In {\em ILP}.

\bibitem[\protect\citeauthoryear{Li, Kambhampati, and Yoon}{2009}]{li2010learning}
Li, N.; Kambhampati, S.; and Yoon, S.
\newblock 2009.
\newblock Learning probabilistic hierarchical task networks to capture user preferences.
\newblock In {\em IJCAI},  1754–1759.

\bibitem[\protect\citeauthoryear{Lotinac and Jonsson}{2016}]{lotinac2016constructing}
Lotinac, D., and Jonsson, A.
\newblock 2016.
\newblock Constructing hierarchical task models using invariance analysis.
\newblock In {\em ECAI}.
\newblock  1274--1282.

\bibitem[\protect\citeauthoryear{Nau \bgroup et al\mbox.\egroup }{1999}]{SHOP}
Nau, D.; Cao, Y.; Lotem, A.; and Munoz-Avila, H.
\newblock 1999.
\newblock Shop: Simple hierarchical ordered planner.
\newblock In {\em IJCAI},  968–973.

\bibitem[\protect\citeauthoryear{Nejati, Langley, and Konik}{2006}]{nejati2006learning}
Nejati, N.; Langley, P.; and Konik, T.
\newblock 2006.
\newblock Learning hierarchical task networks by observation.
\newblock In {\em ICML},  665--672.

\bibitem[\protect\citeauthoryear{Richter and Westphal}{2010}]{richter2010lama}
Richter, S., and Westphal, M.
\newblock 2010.
\newblock The {LAMA} planner: Guiding cost-based anytime planning with landmarks.
\newblock {\em JAIR} 39:127--177.

\bibitem[\protect\citeauthoryear{Segura-Muros, Pérez, and Fernández-Olivares}{2017}]{segura2017learning}
Segura-Muros, J.~A.; Pérez, R.; and Fernández-Olivares, J.
\newblock 2017.
\newblock Learning {HTN} domains using process mining and data mining techniques.
\newblock In {\em {{ICAPS}} Workshop on Generalized Planning}.

\bibitem[\protect\citeauthoryear{Seipp, Torralba, and Hoffmann}{2022}]{seipp-et-al-zenodo2022}
Seipp, J.; Torralba, {\'A}.; and Hoffmann, J.
\newblock 2022.
\newblock {PDDL} generators.
\newblock \url{https://doi.org/10.5281/zenodo.6382173}.

\bibitem[\protect\citeauthoryear{Shrobe}{2021}]{shrobehierarchical}
Shrobe, H.~E.
\newblock 2021.
\newblock Hierarchical problem networks for knowledge-based planning.
\newblock In {\em ACS}.

\bibitem[\protect\citeauthoryear{Xu and Mu{\~n}oz-Avila}{2005}]{xu2005domain}
Xu, K., and Mu{\~n}oz-Avila, H.
\newblock 2005.
\newblock A domain-independent system for case-based task decomposition without domain theories.
\newblock In {\em AAAI},  234--240.

\bibitem[\protect\citeauthoryear{Yang, Pan, and Pan}{2007}]{yang2007learning}
Yang, Q.; Pan, R.; and Pan, S.~J.
\newblock 2007.
\newblock Learning recursive htn-method structures for planning.
\newblock In {\em ICAPS Workshop on AI Planning and Learning}.

\bibitem[\protect\citeauthoryear{Zhuo \bgroup et al\mbox.\egroup }{2009}]{zhuo2009learning}
Zhuo, H.~H.; Hu, D.~H.; Hogg, C.; Yang, Q.; and Munoz-Avila, H.
\newblock 2009.
\newblock Learning {HTN} method preconditions and action models from partial observations.
\newblock In {\em IJCAI},  1804–1809.

\bibitem[\protect\citeauthoryear{Zhuo, Munoz-Avila, and Yang}{2014}]{zhuo2014learning}
Zhuo, H.~H.; Munoz-Avila, H.; and Yang, Q.
\newblock 2014.
\newblock Learning hierarchical task network domains from partially observed plan traces.
\newblock {\em Artificial Intelligence} 212:134--157.

\end{thebibliography}
\bibliographystyle{flairs}

\ifthenelse{\isundefined{\includeappendix}}
{
}
{
    \newpage
    \onecolumn 

\twocolumn[\centering \shortstack{\huge Appendix\\[3ex]}]
\appendix
\section{Landmark Orderings}
\label{sec:landmarks_appendix}
A landmark graph comes with orderings among landmarks.
For a plan of length $n$, a landmark $l_k$ for $0 \leq k \leq n$ occurs from the initial state if $k=0$  and action $a_k$ otherwise; \firstadded{l_k} denotes the first occurrence. 
There are four types of orderings among landmark \leftlm and \rightlm for $0 \leq i < j \leq n$ \cite{richter2010lama}:
\begin{itemize}
    \item A \textit{natural} ordering \leftlm $\rightarrow_{n}$ \rightlm for $i < j$, indicates that \leftlm occurs at some point before \rightlm.
    
    \item A \textit{necessary} ordering \leftlm $\rightarrow_{nec}$ \rightlm for $i = j-1$ indicates that \leftlm occurs one step before \emph{one of} the times \rightlm occurs.  A necessary ordering is a natural ordering.
    
    \item A \textit{greedy-necessary} \leftlm $\rightarrow_{gn}$ \firstrigthlm for $i =j-1$ indicates that \leftlm occurs one step before \emph{the first time} \firstrigthlm occurs. A greedy-necessary ordering is a necessary ordering.
    
    \item A \textit{reasonable} ordering \firstleftlm $\rightarrow_{r}$ \rightlm for $h < i < j$ indicates that \rightlm \emph{reoccurs} some point after \emph{the first time} \firstleftlm occurs; i.e., \rightlm previously occurred at some time $h$ and became false at $h+1$.   This ordering may save effort in solving the problem, thus it is "reasonable".
    
\end{itemize}

\section{Detailed Experimental Methodology}
\label{sec:domain_description_appendix}

We tested \curriculama against \htnmaker in the following five domains:
\begin{itemize}
\item
The \emph{Blocks World} encompasses scenarios where a number of blocks are arranged on a table, potentially stacked upon each other, with a robotic hand tasked with reorganizing them. Our study's focus in this domain is on the reorganization of 5 blocks.
\item
In the \emph{Logistics} domain, the primary task involves package delivery within and between cities, utilizing trucks and airplanes. A typical scenario in our study involves the delivery of one package across 3 cities, each having 2 specific locations.
\item
The \emph{Rover} domain involves navigation through waypoints and sample collection tasks of objectives on a planetary surface. We study problems that have at most 2 rovers, 5 waypoints, and 6 objectives.
\item
In the \emph{Satellite} domain, the objective is to collect various images using satellites equipped with different observation instruments.  We study problems that have at most 4 satellites, 3 instruments, 8 instrument modes, and 6 targets.
\item
\emph{Zeno Travel} involves transportation of passengers between cities using aircraft with varying fuel levels, which are represented numerically using predicate logic. We study problems that have at most 6 cities, 5 aircraft, and 5 passengers.
\end{itemize}

\noindent
We repeated  five trials in each of the five domains and reported on the following metrics:

\begin{itemize}
    \item \emph{Convergence}: We measured the convergence rate of the proportion of test problems the planner could successfully solve with the methods learned by each HTN learner. If the methods learned from only a few examples are sufficient to solve most of the testing problems, we say that the set of methods rapidly converges to one that is complete.
    \item \emph{Average Plan Length}: We measured the average length of the plans for the successfully solved test problems, which informs us about the efficiency and complexity of the solutions generated by the learning algorithms.
    \item \emph{Average Planning time}: We measured the average amount of time that the planner needed to solve the test problems using the learned methods.
    \item \emph{Method Generation}: We recorded the cumulative number of methods generated by each of the HTN learners as they saw more and more training examples.
    \item \emph{Running Time}: We compared the running time of the HTN learners at different stages of the learning process.
\end{itemize}

\section{More Experiments}
\label{sec:appendix:more_exp}
On the next two pages are figures showing the experimental results for the five domains in our experiments: Blocks World, Logistics, Satellite, Rover, and Zeno Travel.
Figure \ref{fig:exp_planning} shows the convergence analysis, Figure \ref{fig:exp_planning_length} shows the average plan lengths, Figure \ref{fig:exp_planning_time} shows the time to solve planning problems using the learned methods, Figure \ref{fig:algorithm_times} shows the running time needed to learn methods, and Figure \ref{fig:exp_planning_number_methods} shows the number of methods learned.
\begin{figure*}[ht]
\includegraphics[width=\textwidth]{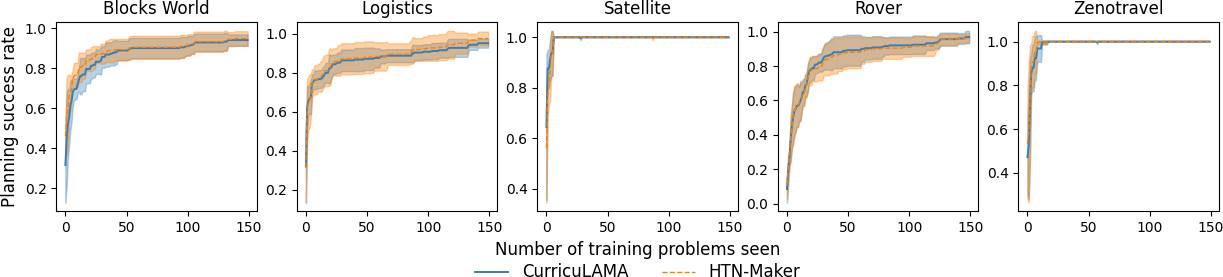}
\caption{
\emph{Convergence analysis.}
The y-axis shows the fraction of problems that the planner could successfully solve using the methods that each learning algorithm learned, and the x-axis shows the number of training problems from which the methods were learned.
The shaded regions show the variance in problems solved across five separate trials.
}
\label{fig:exp_planning}
\end{figure*}

\begin{figure*}[ht]
\centering
\includegraphics[width=\textwidth]{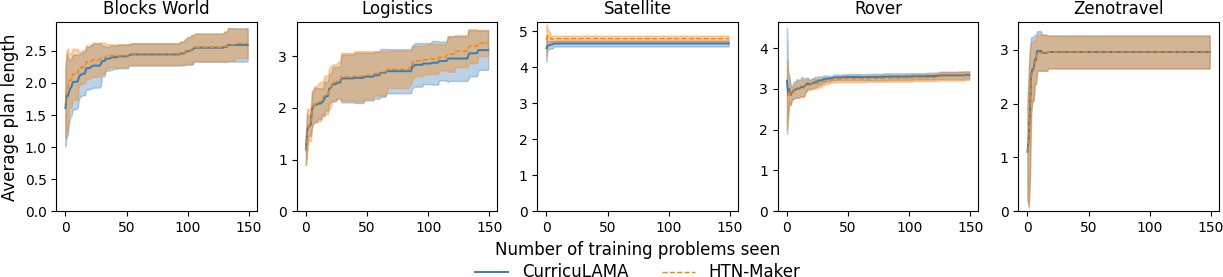}
\caption{
\emph{Average plan lengths.}
The y-axis shows the average length of the plans that the planner produced using the learned methods,
and the x-axis shows the number of training problems from which the methods were learned, ranging from zero to 150 training problems.
The shaded regions show the variance in plan length across five separate trials.
}
\label{fig:exp_planning_length}
\end{figure*}

\begin{figure*}[ht]
\centering
\includegraphics[width=\textwidth]{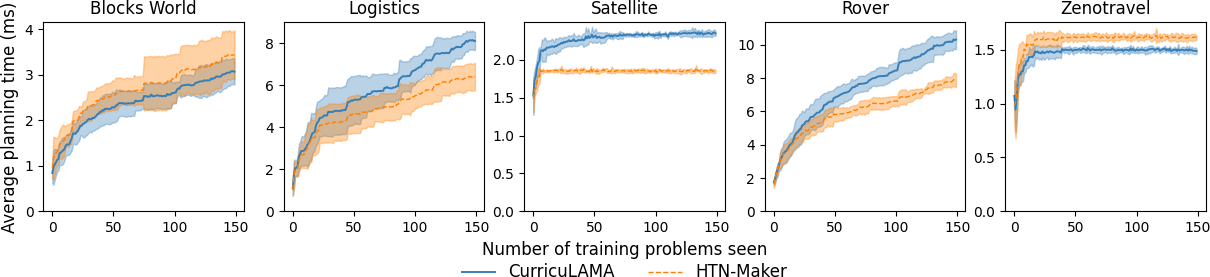}
\caption{
\emph{Time to solve planning problems using the learned methods.}
The x-axis gives the number of training problems from which each learning algorithm learned its methods, and the y-axis gives the average planning time over the 50 test problems. The shaded regions denote the variance in planning time across five separate trials.
}
\label{fig:exp_planning_time}
\end{figure*}

\begin{figure*}[ht]
\centering
\includegraphics[width=\textwidth]{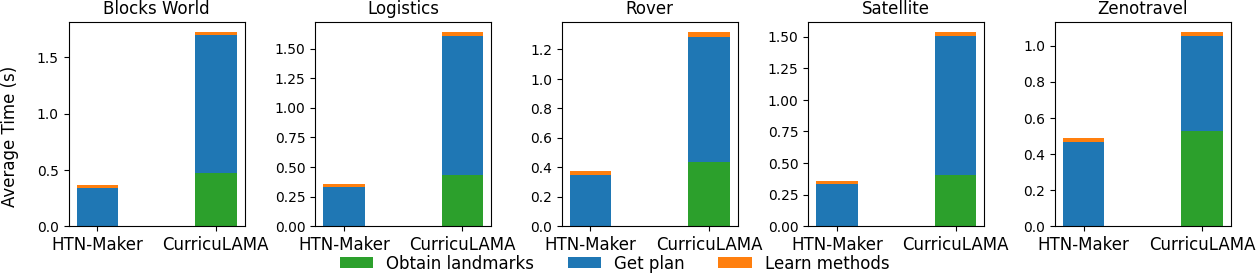}
\caption{
\emph{Running time needed to learn methods.}
 The bars represent the average time that each learning algorithm spent on different parts of the learning process. Green represents the time to obtain landmarks, blue indicates the time to get the plan, and orange shows the time to learn methods.
 }
\label{fig:algorithm_times}
\end{figure*}

\begin{figure*}[ht]
\centering
\includegraphics[width=\textwidth]{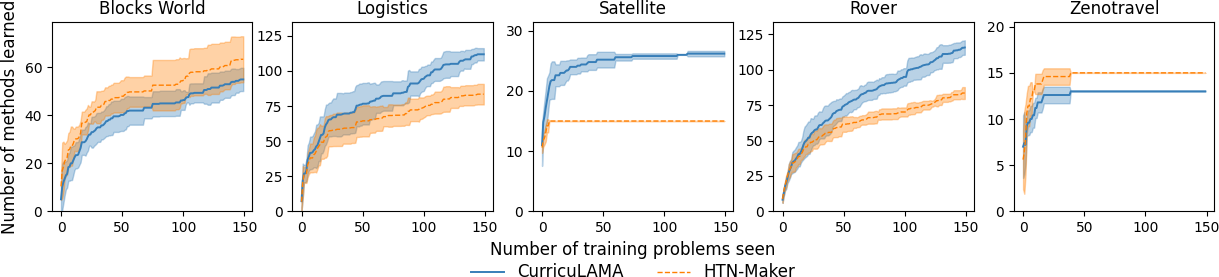}
\caption{\emph{Number of methods learned.} 
The y-axis is the total number of methods learned, and the x-axis is the number of training problems from which they were learned. Both algorithms show increases in the number of methods as they are exposed to more training problems. The shaded areas indicate the variance in the number of methods learned across five trials.
}
\label{fig:exp_planning_number_methods}
\end{figure*}

\section{Landmarks May Produce Suboptimal Subgoal Ordering}
\label{sec:illustration_appendix}
\begin{figure*}[h]
\centering
\includegraphics[width=.45\textwidth]{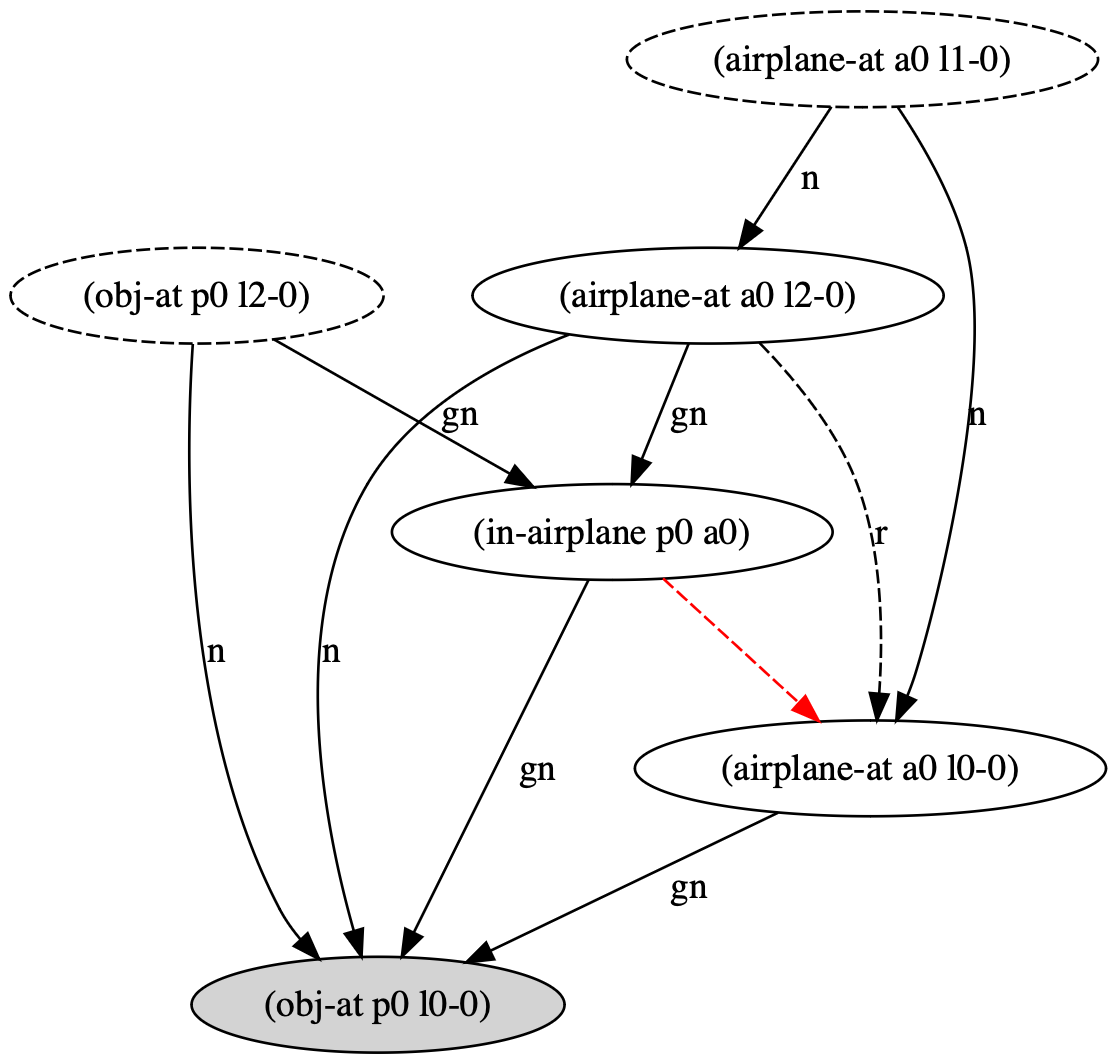}
\caption{Landmark graph illustrating potential suboptimal planning in \curriculama. Landmarks are represented by circles, and edges indicate ordering: `r' for reasonable, `n' for natural, and `gn' for greedy necessary. A missing reasonable ordering would prioritize \code{(in-airplane p0 a0)} before \code{(airplane-at a0 l0-0)} to avoid unnecessary airplane movements (marked by the red dashed arrow).}
\label{fig:lm_graph}
\end{figure*}
\curriculama's planning mechanism may cause it to learn extraneous methods in some domains.
For illustrative purposes, consider delivering a package p0 from location l2-0 to location l0-0 in the Logistics domain.
The airplane a0 is initially at location l1-0.
The most efficient strategy after flying the airplane a0 from l1-0 to l2-0 (according to the first landmark \code{(airplane-at a0 l2-0)}) would be to load the package into the airplane followed by a direct flight to l0-0. Nonetheless, as depicted in Figure~\ref{fig:lm_graph}, \curriculama may adopt a sequence where the airplane first relocates to l0-0 without the package, resulting in a suboptimal path and thus an increase in the number of methods learned.

}

\end{document}